\def\year{2022}\relax
\documentclass[letterpaper]{article} 
\usepackage{aaai22}  
\usepackage{times}  
\usepackage{helvet}  
\usepackage{courier}  
\usepackage[hyphens]{url}  
\usepackage{graphicx} 
\usepackage{natbib}  
\usepackage{caption} 
\DeclareCaptionStyle{ruled}{labelfont=normalfont,labelsep=colon,strut=off} 
\usepackage{comment}
\usepackage{graphicx}

\usepackage{textcomp}
\usepackage{xcolor}
\def\BibTeX{{\rm B\kern-.05em{\sc i\kern-.025em b}\kern-.08em
    T\kern-.1667em\lower.7ex\hbox{E}\kern-.125emX}}
\newcommand{\RebeccaNote}[1]{$\ll$\textcolor{purple}{Rebecca}$\gg$}
\usepackage{soul}
\usepackage{subcaption}

\usepackage[T1]{fontenc} 
\usepackage{setspace} 
\usepackage{dialogue}
\usepackage{calc} 

\newlength{\widestname}
\makeatletter
\renewenvironment{dialogue} {%
    \begin{list}{} {%
        \setlength\itemsep{\z@ \@plus .5ex} %
        \setlength{\parsep}{\parskip} %
        \setlength{\rightmargin}{0pt} 
        \setlength{\labelwidth}{\widestname} 
        \setlength{\labelsep}{0.5em} 
        \setlength{\leftmargin}{\labelwidth} 
        \addtolength{\leftmargin}{\labelsep} 
        \defcommand\speak [1] {\item[{##1}]} 
        
      }%
      \PreDialogue\relax
    }{%
  \end{list}%
  }
\makeatother

\usepackage[utf8]{inputenc}
\usepackage{amsmath}
\usepackage[ruled,vlined, linesnumbered]{algorithm2e}

\RestyleAlgo{ruled}
\SetKwComment{Comment}{/* }{ */}

\title{A Grounded Observer Framework for Establishing Guardrails for Foundation Models in Socially Sensitive Domains}

\author{
    Rebecca Ramnauth$^{1}$, Dra\v{z}en Br\v{s}\v{c}i\'{c}$^{2}$, Brian Scassellati$^{1}$
}
\affiliations{
    \textsuperscript{1} Department of Computer Science, Yale University, New Haven, CT, USA \\
    \textsuperscript{2} Graduate School of Informatics, Kyoto University, Kyoto, Japan

    rebecca.ramnauth@yale.edu
}

\usepackage{bibentry}

\begin{document}

\maketitle

\begin{abstract}
As foundation models increasingly permeate sensitive domains such as healthcare, finance, and mental health, ensuring their behavior meets desired outcomes and social expectations becomes critical. Given the complexities of these high-dimensional models, traditional techniques for constraining agent behavior, which typically rely on low-dimensional, discrete state and action spaces, cannot be directly applied. Drawing inspiration from robotic action selection techniques, we propose the grounded observer framework for constraining foundation model behavior that offers both behavioral guarantees and real-time variability. This method leverages real-time assessment of low-level behavioral characteristics to dynamically adjust model actions and provide contextual feedback. To demonstrate this, we develop a system capable of sustaining contextually appropriate, casual conversations (``small talk''), which we then apply to a robot for novel, unscripted interactions with humans. Finally, we discuss potential applications of the framework for other social contexts and areas for further research.
\end{abstract}

\section{Introduction}

Foundation models are rapidly being integrated into various fields, from medical diagnostics and financial predictions to socially sensitive areas such as education, mental healthcare, and support for individuals with disabilities. Despite being aware of the inherent risks of AI hallucinations, misinformation, and bias, a recent large-scale global study revealed that 66\% of respondents are still willing to use this nascent technology in sensitive areas such as personal advice and relationship counseling \cite{capgemini_ai}. This paradox highlights the immense potential benefits of these models in addressing societal challenges while also underscoring the current concerns. A significant issue tempers the widespread adoption of these tools: the lack of comprehensive guardrails to prevent undesired behavior and ensure reliable outcomes. 



In fields where accuracy and reliability are paramount, such as healthcare and finance, the consequences of errors can be severe. Yet, in socially sensitive domains, where the parameters of success are less tangible, the impact of missteps can be as profound. For example, a system intended to provide calming techniques in a clinic waiting room could exacerbate anxiety if it delivers generic or poorly timed suggestions. If it fails to recognize the urgency or context of a patient's distress, it may offer advice that feels dismissive or irrelevant, potentially increasing the patient's anxiety. In light of such effects, foundation models should have robust guardrails to protect users and the system's integrity. 

Designing usable systems that impose limits on foundation models involves two key challenges. First, foundation models are based on statistical learning from vast datasets, making their internal mechanisms complex and opaque. Traditional rule-based systems use symbolic representations, which are formal and interpretable but not directly compatible with the statistical nature of foundation models. This difficulty is compounded when integrating symbolic rule-based systems that map human concepts into precise rules, a challenge akin to reconciling statistical learning mechanisms with symbolic representation systems. While neurosymbolic approaches that aim to blend statistical and symbolic methods are being explored (e.g., \citealp[]{garcez2023neurosymbolic}), effective integration remains an open area of research.

Second, foundation models must be able to adapt their behavior in real-time to the unique needs and contexts of individual users \cite{wang2023adapting, chen2024large}. Static, predefined rules often do not address the dynamic and nuanced nature of personal interactions \cite{raman2022planning}. For instance, a large language model (LLM) for mental health support must respond appropriately to a user's current emotional state and context. A static rule-based approach may fail to provide suitable support during a crisis or tailor interactions based on ongoing conversations, highlighting the need for real-time adaptability to meet individual user needs.

These two challenges are not unique to foundation models but manifest in other areas, such as robotics. In action selection for robot systems, an agent must decide on actions to take, often using large-scale statistical models, while adhering to user-specified rules, such as ``don't touch the stove.'' Addressing this involves techniques known as shielding \cite{alshiekh2018safe} and interactive policy shaping \cite{griffith2013policy}. Shielding techniques prevent particular actions from being executed, effectively restricting the robot's behavior, while interactive policy shaping modifies the action selection policy in real time based on user input or situational changes. These approaches aim to reconcile the flexibility of statistical models with the necessity of adhering to predefined constraints \cite{biza2021action}, reflecting similar challenges faced in the context of foundation models.

Drawing inspiration from robotic action selection techniques, we propose a framework for constraining foundation model behavior that offers both behavioral guarantees and real-time variability. This method involves a grounded observer that continuously assesses the underlying model's candidate actions based on low-level behavioral characteristics, makes dynamic adjustments to the model's action generation, and provides feedback directives to ensure the behavior remains contextually appropriate and effective.

In this paper, we present the conceptual framework of the grounded observer for establishing guardrails for foundation models. We apply this framework to build agents capable of small talk, a task that requires nuanced social sensitivity to ensure continued appropriateness and relevance. This case study of small talk demonstrates how the grounded observer can impose precise constraints on LLM behavior in highly subjective contexts and challenge the typically informative and assistive nature of these models. We also demonstrate that this method leads to more positive and socially appropriate interactions when integrated into a robot where its embodiment amplifies social impacts. Lastly, beyond small talk, we explore how this technique can be applied to create guidelines in various socially sensitive domains.

\section{Related Work}
Given their complexity and the vast datasets they are trained on, ensuring that foundation models behave in predictable and socially acceptable ways is a significant challenge. Researchers have explored approaches to impose constraints on these models, each with strengths and limitations.



\subsection{Prompt Engineering}
The current standard for constraining model behavior is having a good prompt. While crafting specific input prompts has shown promise in many applications \cite{giray2023prompt, mesko2023prompt, white2023prompt}, it has significant limitations when it comes to robustly constraining agent behaviors, especially in complex, dynamic, and sensitive contexts. 

\emph{Lack of Robustness.} One of the primary limitations of prompt engineering is its lack of robustness. While specific prompts can guide the model in controlled scenarios, they often fail to generalize across different contexts and variations. A prompt that works well in one situation might produce unexpected or undesirable results in another, leading to inconsistent behavior \cite{zhou2022large, huang2024selective}.

\emph{Context Sensitivity.} Foundation models are highly sensitive to the context provided by prompts. Small changes in phrasing can lead to significantly different outputs, making it challenging to predict and control the model's behavior reliably \cite{denny2023conversing, dong2024building}. This sensitivity can be particularly problematic in dynamic environments where the context is continuously changing.

\emph{Inability to Enforce Hard Constraints.} Prompt engineering cannot enforce hard constraints on model behavior. While prompts can suggest or guide the model toward certain behaviors, they cannot guarantee that it will always comply with these suggestions \cite{niknazar2024building}. This limitation is critical in applications where strict adherence to ethical guidelines or safety protocols is necessary.

\emph{Translating to Real-World Behavior.} Many real-world scenarios involve ambiguous and complex situations that are difficult to capture with prompts \cite{leite2013social}. For instance, ensuring that an LLM provides appropriate mental health support requires understanding and responding to nuanced emotional cues, which cannot be fully encapsulated in a prompt. In such cases, prompt engineering alone cannot ensure reliable and sensitive behavior.

\emph{Temporal Constraints.} Prompt engineering does not inherently support temporal constraints, where the desired behavior depends on the sequence and timing of interactions \cite{lyu2024keeping, chen2023forgetful}. For example, maintaining consistent behavior over multiple exchanges with a user is challenging to achieve through prompt design alone.



\subsection{Constrained Reinforcement Learning}

Constrained reinforcement learning (CRL) enhances traditional RL by integrating predefined constraints to ensure agents operate within specific safety, ethical, or operational boundaries. While traditional RL focuses solely on maximizing cumulative rewards, CRL incorporates additional constraints as hard limits (e.g., avoiding unsafe actions) or soft constraints (e.g., minimizing deviation from desired behaviors). CRL incorporates inductive biases through logical rules that govern the agent's behavior, applying these constraints directly to states and actions or modifying the reward function to align with the defined limits \cite{gu2022review}.

A notable approach within CRL is shielded RL, which employs user-defined policy overrides, or ``shields,'' to restrict certain actions based on specific conditions, thereby minimally disrupting the RL model while enforcing desired behaviors \cite{garcia2015comprehensive}. However, shielded RL typically relies on a dynamic model and repairing existing policies rather than adapting to evolving preferences. In contexts such as personalized healthcare or companionship, a flexible approach to adapt policies to meet context-specific needs in real-time is more suitable. 

\subsection{Transparent Matrix Overlays}
Transparent Matrix Overlays (TMOs) is a promising technique for real-time modification of agent behavior by integrating user directives as symbolic constraints on a robot's policy \cite{brawer2023interactive}. This approach merges concepts from CRL and shielded RL, leveraging symbolic reasoning to enhance flexibility in behavioral adaptation. 

Demonstrated through a simulated collaborative cooking task \cite{brawer2023interactive}, TMOs allowed adjustments to a robot's policy without requiring extensive retraining. By applying logical rules and user-specific directives as temporary constraints, TMOs facilitated immediate changes in behavior to align with evolving user preferences. This method contrasts with traditional CRL techniques, which often require substantial retraining to incorporate new constraints, and shielded RL methods that focus on policy repairs rather than accommodating real-time preference changes. This approach balances the stability of learned behaviors with the flexibility required to meet new and evolving preferences, making it a valuable tool for interactive systems.

One limitation of TMOs is the reliance on hand-crafted predicates and classifiers. In the current implementation, these elements are manually designed to define constraints and directives. While this method works within controlled environments, it constrains the flexibility of the TMO approach. The assumptions of having a relatively simple, discrete state space, deterministic actions, and non-parallel task completion further simplify the scenario. Real-world applications often involve more dynamic and complex environments where these assumptions may not hold.

\subsection{State and Action Space Abstraction} 

Most action selection mechanisms, like TMOs, assume a known, discrete, or discretized state space with well-defined actions. However, for foundation models, an action selection mechanism must handle continuous and possibly infinite state spaces where iterating through all possible actions or states may be impractical \cite{paul2024continually}. This requires rules that can overlay abstracted state representations or symbolic predicates to approximate the agent's internal state and action space. Instead of exhaustively evaluating every action, the agent can use these overlays to focus on a manageable subset of candidate actions or employ probabilistic sampling techniques within the space emphasized by the overlays.
Furthermore, 
such abstraction must supersede differences in how proprietary architectures handle context, manage memory, and generate responses \cite{naveed2023comprehensive}.



\section{The Grounded Observer Framework}
Social behavior is inherently emergent and complex. However, in many cases, appropriate behavior can be guided by simple rules. 
Just as TMOs embed rules to control behavior, we can apply similar principles to ensure that foundation models exhibit appropriate social behavior. Foundation models are analogous to the action policies generated---they are statistical models that are expensive to generate, difficult to dissect, and opaque to inspection. By imposing transparent and adaptive constraints, we can manage and direct these models to align with desired outcomes in socially sensitive domains. This can be achieved by evaluating a model's output through context-based rules and providing feedback to guide the model toward more appropriate behaviors.

\subsection{Overview of the Framework}

\begin{figure}[t]
    \centering
    \includegraphics[width=\columnwidth]{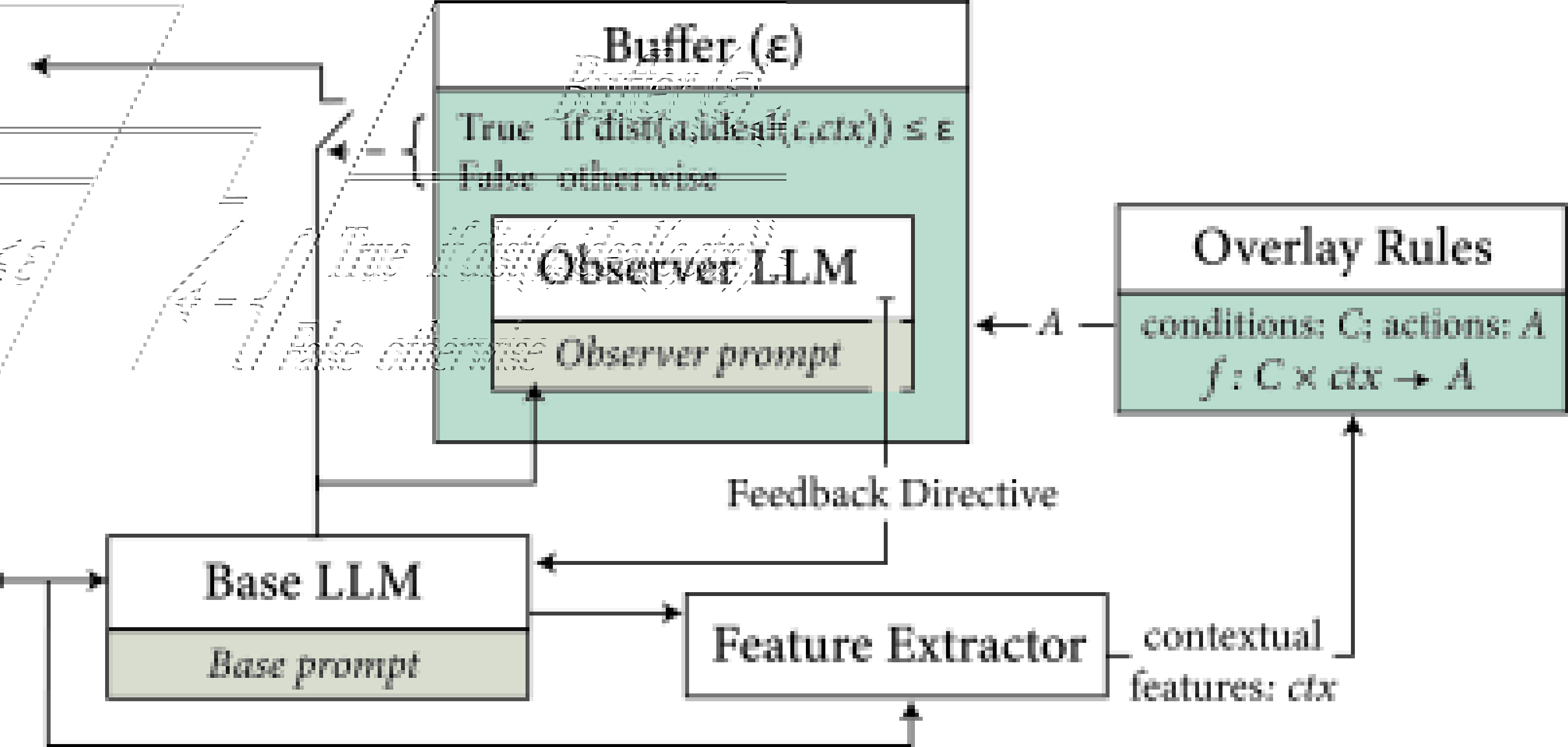}
    \caption{The grounded observer monitors a base model's behavior to ensure responses adhere to overlay constraints.} 
    \label{fig:observer}
\end{figure}

We begin with a foundation model, referred to as the \emph{base} model in Fig \ref{fig:observer}, which generates actions in response to environmental or user inputs. Depending on the type of model, these actions can take the form of text, images, or other outputs. To provide a clear overview in this section, we will focus on LLMs, assuming that both the model's inputs and outputs are in text form, though other modalities are also applicable. To evaluate the base model's actions, feature extractors convert these actions and the surrounding context into numerical features. These features can then be analyzed as scores based on the characteristics we want to evaluate.  Depending on the scenario, these extractors may also incorporate inputs from high-level planners or context observers. For example, a feature extractor could be designed to quantify the politeness of the model's text output. 

These contextual features are evaluated against IFTTT (If This, Then That) rules, which function as overlays on the model's actions. Think of these rules as semi-transparent sheets on an overhead projector: you can stack, prioritize, or remove them to adjust the view without altering the original image. Similarly, these rules can be adjusted without extensive changes to the base model. 

High-level descriptors---summaries of how well proposed actions align with the overlays---are given by each overlay rule in a fixed text structure. These descriptors pinpoint areas where proposed actions comply with or deviate from the established rules. For instance, a rule about politeness might provide a directive like ``tone is too polite,'' while a rule that assesses user frustration could direct the model to include more empathetic language. Each overlay also produces a score indicating the degree of deviation from the rule. These scores highlight more severe rule violations by using methods such as ranking or incorporating keywords like ``prioritize'' or ``urgent'' in the directives. 


An \emph{observer}, a separate foundation model instance, receives these directives, then combines and translates them into actionable feedback for the base model. For example, if a directive indicates that the tone is too polite, the feedback might be, ``The previous response was overly formal. Please adopt a more casual tone.''

\subsection{Action Filtering} 
A buffer acts as a gatekeeper, as shown in Fig. \ref{fig:observer}, determining whether a proposed action should be accepted. Each overlay can be assigned a rigidity parameter (depicted as $\epsilon$) that defines how strictly the model must adhere to the rule. Essentially, in reference to the overhead projector analogy, this parameter controls the translucency of an overlay. Instead of enforcing a strict binary compliance---where actions either fully meet the overlays or not---rigidity offers a gradient of compliance or a buffer around proposed actions.

For highly rigid overlays, compliance is strictly enforced. If an action or response deviates from the specified rules, the base model is required to regenerate new candidate actions. This ensures that only actions meeting the strict criteria are considered. For instance, if an overlay rule demands that responses must be empathetic, any response lacking empathy would lead to the base model generating alternative responses that conform to this requirement.

Less rigid overlays allow the overlay to be more translucent so that actions that partially meet the criteria can still be considered. The observer model may rank or prioritize these partially compliant actions, accepting them within a permissible range. For example, if an overlay requires responses to be empathetic, a response that shows limited empathy but is otherwise acceptable might still be chosen.

This flexibility helps manage the model's load and processing time when correcting its actions. For non-critical conditions, low rigidity can be used, while critical conditions require higher rigidity. The buffer can limit the number of action regeneration cycles to prevent excessive resource consumption while enforcing the necessary constraints.

\subsection{Feedback Directives}
The observer utilizes the overlay descriptors and rigidity to create targeted feedback prompts to the base model. We incorporate two types of feedback:




\emph{Implicit feedback} notes that the action is acceptable but offers constructive advice for improving subsequent actions. For example, if the actions are near compliance but not perfect, implicit feedback may recommend minor adjustments, such as modifying tone or phrasing. Suppose the base model generates a response that is mostly empathetic but could be softer in tone. The implicit feedback might suggest: ``Consider using a gentler tone in your responses.'' This allows the base model to refine its output in future iterations.

\emph{Forced feedback} is employed when the base model’s actions significantly deviate from the overlay constraints. When the descriptors reveal substantial misalignment with the overlay rules, the observer generates a more directive prompt, instructing the base model to focus on specific improvements until it fully complies with the constraints. The observer may issue several rounds of feedback if needed until proposed actions meet the overlay requirements. 

Overall, this feedback loop ensures that the base model continually aligns with the overlays by translating its performance on specific rules into clear instructions. In the next section, we apply this framework and demonstrate the role of each component within a social context.

\section{Creating Agents Capable of Good Small Talk}
Imagine a modern care home for the elderly where a state-of-the-art robotic assistant, designed to enhance residents' well-being, manages routine healthcare tasks. Alex, a resident, seeks a connection beyond the daily routine and attempts to chat with the robot:
\begin{dialogue}
    \speak{Alex} Hi CareBot, how's it going?
    \speak{Bot} Hello. How may I help you?
    \speak{Alex} Oh, just making conversation. Anything interesting happen in your world?
    \speak{Bot} I have access to a vast database of news articles. Would you like information on a specific topic? 
    \speak{Alex} No, never mind that. The weather will be nice this weekend. How would you spend it?
    \speak{Bot} The weather forecast expects daytime highs around 75°F and comfortable evening lows of 60°F... 
\end{dialogue}

Today, an essential component of designing intelligent systems is to imbue some level of speech, language understanding, and conversational behavior \cite{shieber2004turing, fu2022learning}. Despite the potential for these intelligent agents to elicit meaningful interactions, the dialogue between Alex and the robot exemplifies a common shortcoming. Alex initiates a friendly exchange, expressing a desire for casual conversation with the robotic assistant. However, the robot, proficient in providing information, struggles to reciprocate the informal nature of the dialogue. Instead, the robot redirects the conversation towards its programmed functionalities, offering information and task-oriented assistance. 

Although the boundaries of types of conversation are always uncertain, ``small talk'' has a recognized currency in several traditions of sociolinguistics and communication studies \cite{coupland2014small}. 
It can be defined as a generally informal and light-hearted conversation with a social purpose aimed at building or sustaining interpersonal connections rather than conveying substantial information. 
Yet, small talk does not have a strict formula, as it is inherently flexible and context-dependent. This fluid nature presents a significant challenge for current-day LLMs, which often rely on structured and well-defined question-answer patterns.

Yet, the literature emphasizes distinct characteristics of small talk \cite{laver1981linguistic, eggins2004analysing}. One key aspect is \emph{brevity}, where responses are typically concise, avoiding unnecessary elaboration or verbosity. Another essential characteristic is \emph{tone}; responses maintain a light and informal tone, steering clear of negativity, complaints, or contentious topics. \emph{Non-specificity} is a hallmark of small talk, as it revolves around broad, accessible topics, deliberately avoiding highly specific details. Finally, despite its non-specific nature, small talk maintains \emph{thematic coherence}, staying contextually relevant and focusing on related topics or themes to avoid disjointed elements. The delicate balance among these characteristics highlights both the nuances and the fundamental principles of effective small talk. 

A skilled conversationalist not only learns their partner's preferences over time but also adapts to them in real-time, using naturalistic cues that may be linguistic, implicit, and contextual. For intelligent agents, this means they must swiftly adjust their policies in response to high-level, imprecise, or evolving directives conveyed through natural language. Therefore, we present a proof-of-concept case study of how a grounded observer can dynamically shape an agent's behavior while adhering to high-level directives in a highly subjective social context.

\subsection{Current Landscape of LLM Small Talk}
To establish the baseline of which small talk remains a challenge, we conducted an initial study.\footnote{All methods were preregistered and received IRB approval \cite{ramnauth_brscic_scassellati_2024}.} Three volunteers engaged in 50 conversations each with three distinct state-of-the-art LLMs. Each model had the initial system prompt describing the role as a ``friendly companion who engages in casual, small talk'', with the prior listed criteria definitions. The selected LLMs are GPT-3.5 \cite{brown2020language}, 
for its large-scale language generation capabilities, Gemini Pro \cite{team2023gemini},
for its context-aware bidirectional approach, and LLaMA-2 \cite{touvron2023llama}, 
an autoregressive transformer model fine-tuned on prompt-response pairs. 

\textbf{Data Collection}. The order in which the participants used the LLMs was randomized to mitigate potential order effects. Additionally, conversations lasted at least ten turns, and the interactions occurred over 15 days to allow for conversational variability. The participants engaged with each LLM through a command line interface, unaware of the LLM's name to prevent bias from prior knowledge or familiarity. Following each conversation, assistants rated the ease of each conversation and provided open-ended feedback. Additionally, two research assistants annotated the dataset. Raters were blind to the response speaker and evaluated responses based on recognized small talk criteria: brevity, tone, specificity, and coherence on 5-point Likert scales.  

\label{ref:results-1}

A total of $150$ conversations were transcribed, yielding an average of $10.31$ responses per conversation ($SD = 1.13$). This led to a total of $1547$ annotated responses. Due to the inherent ambiguity of criteria evaluation, we calculated the inter-rater reliability for a randomly selected subset of 20 conversations, constituting $13.3\%$ of the total dataset. 
Inter-rater reliability was calculated using contingency tables, employing Cohen's Kappa ($\kappa$), with the observed agreement and the distribution of ratings for each rater. 
The resulting $\kappa$ values were $0.81$ for brevity, $0.78$ for tone, $0.74$ for specificity, and $0.65$ for coherence.

\begin{figure}[t]
    \centering
    \includegraphics[width=\columnwidth]{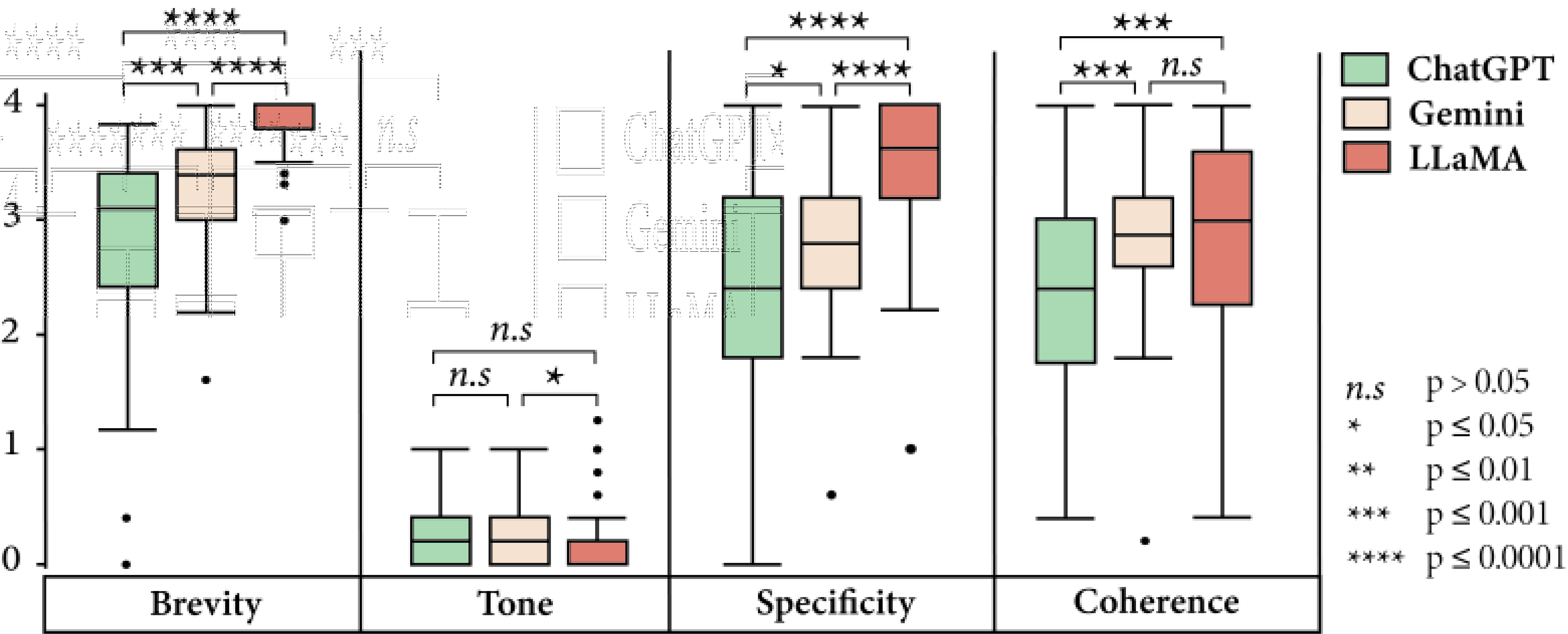}
    \caption{Evaluation Scores of LLMs. This graph reflects the similarity of the model's small talk to that of the participants, scored from 0 (no difference between human and model responses) to 4 (highest absolute difference).}
    \label{fig:LLM-human-likeness}
\end{figure}

\textbf{Human vs. Agent Comparison.} 
We used paired dependent t-tests to assess the differences between the agents' and humans' responses across the small talk criteria. 
A conventional $\alpha$ of $0.05$ was employed, and resulting p-values were Holm-corrected to control the familywise error rate. 
The results revealed significant differences in brevity ($t = 86.78$, $p \leq 0.0001$), 
tone ($t = 1.70$, $p = 0.04$), 
specificity ($t = 58.06$, $p \leq 0.0001$), 
and thematic coherence ($t = -55.72$, $p \leq 0.0001$) between the agent 
and human responses. 
This suggests that LLM-generated small talk responses were notably less concise, somewhat more positive, more specific, and less thematically coherent compared to human responses. 
We summarize the degree of similarity between LLM behavior and human responses by computing the absolute difference in their average scores across these dimensions within each conversation. The ``human-likeness'' of each LLM is illustrated in Fig. \ref{fig:LLM-human-likeness}, where $0$ represents no difference at all and $4$ is the highest absolute difference between human and LLM responses. 



\textbf{Impact of Forgetfulness.} We explored whether LLMs' low performance in small talk results from ``forgetfulness'' of the initial prompt by using mixed-effects modeling. This model analyzed the relationship between the response sequence index and our outcome variables, accounting for the conversation identifier and LLM name as random effects to address the nested data structure.

For brevity, a significant positive coefficient ($\beta$ = 0.10, $p \leq 0.001$) indicated increased wordiness of the agents' responses as the conversation progressed. Specificity showed a significant positive association ($\beta$ = 0.11, $p \leq 0.001$), indicating the agents' responses become more specific over time. Coherence showed a significant negative coefficient ($\beta$ = -0.10, $p \leq 0.001$), suggesting the agents became less coherent over time. Tone did not exhibit a significant relationship with the response index ($\beta$ = 0.00, $p > 0.05$). 

\textbf{Feedback}. 
Open-ended feedback highlighted participants' difficulties in conversing with LLMs, which we categorized into four themes through informal thematic analysis.
Often, conversations (59\%, $N=89$ out of 150 conversations) ended abruptly or felt forced, with one user commenting, ``The bot didn't encourage more conversation than I expected. I'm not sure how to continue in a way that doesn't feel forced.'' Additionally, 51\% ($N=77$) of conversations featured multiple questions or rapid topic shifts, leading to confusion. One participant noted, ``It was hard to follow because the bot asked so many questions and touched upon so many different topics in the same response.'' Emotional loops affected 23\% ($N=34$) of conversations, where LLMs intensified emotional aspects without appropriate transitions. As one user stated, ``I felt that the bot was leading the conversation down a rabbit hole.'' Finally, 68\% ($N=102$) of conversations involved excessive advice or detailed information, which felt like reprimands rather than balanced dialogue. A user remarked, ``I felt I was reprimanded for conveying an opinion.'' These issues highlight the need for strategies to ensure small talk interactions are coherent, balanced, and contextually appropriate.

\subsection{Observer-Enabled Small Talk}
It is evident from the initial study that there is a disparity in how LLMs maintain conversational momentum versus what is expected or exhibited by human speakers. Building on these insights, we apply the grounded observer framework to develop agents adept at sustaining small talk. We employ two instances of GPT-3.5, one as the base model and the other as the observer, because it performed relatively well (Fig. \ref{fig:LLM-human-likeness}). By using the same base model prompt, we can compare the performance of an observer-enabled system against the baseline results, assessing how improvements can be achieved despite the same base model configuration. 

To design the overlay rules, we extract specific features based on response criteria emphasized in the literature: brevity, tone, specificity, and coherence. We estimate the rigidity and thresholds for the overlays using the dataset collected from the baseline study. Below, we describe the methods for calculating these features, followed by a description of the feedback prompts generated by the observer.




\textbf{Brevity}. Setting a limit on the length of the generated responses enhances the practicality and user-friendliness of the model, aligning with the natural flow of everyday conversations. To enforce this limit, the observer module defines an expected number of completion tokens \cite{openai_chatgpt}. 

\textbf{Tone}. We employed the VADER model \cite{hutto2014vader} for sentiment analysis. The evaluation of tone and sentiment in a small talk response can be approached both per sentence and holistically. This dual approach provides a nuanced understanding of the response's contribution to the conversational tone, addressing micro-level details and macro-level coherence. We estimated the relative weights of these scores using the baseline dataset and calculated a combined score (C) as follows: 
\[
C = H \times w_H + \frac{1}{n} \sum_{i=1}^{n} s_i \times w_i
\]
In this formula, \( H \) represents the overall score from VADER, and \( w_H \) is the weight assigned to this overall score. The variable \( n \) denotes the number of sentences, while \( s_i \) indicates the sentence-level score for the \( i \)-th sentence, with \( w_i \) being the weight assigned to that specific sentence.
The score $C$ ranges from $-1$ to $+1$. A value between $-0.5$ and $0$ signifies a neutral response, and from $0$ to $1$ indicates positivity---both are acceptable for a small talk response. 

\textbf{Specificity}. Response specificity is assessed through NLTK's named entity chunker and part-of-speech tagging \cite{bird2009natural}. Counts of entities and descriptive words are normalized based on maximum expected counts, derived from human responses in the baseline data

\textbf{Coherence}. To quantify coherence, we encoded each response into a sequence of tokens and derived embeddings using BERT \cite{devlin2018bert}. The calculated entropy of token embeddings of a response captures the uncertainty and diversity at each conversational turn. Subsequently, we gauged information gain by considering the entropy of the previous response and the weighted average of the entropies in the current response.

\textbf{Other Considerations}. As noted in baseline study, it is the nature of LLMs to offer assistance.
Yet, offers of help may result in conversations that sound too technical or formal. To mitigate this, the observer calculates the cosine similarity of embeddings to keywords of assistance, such as ``help'', ``assist'', and ``information''. We determined the list of specified keywords using the collected dataset. 

\textbf{Feedback}. Timely responses are crucial for maintaining conversational flow, which requires balancing the detail and frequency of model updates during execution. When the base model generates a response that violates an overlay rule, such as being excessively verbose, the permissible buffer allows a gradation of compliance. For minor deviations, the buffer will allow the observer to synthesize the overlay directives to curate implicit feedback such as, ``Your response was too lengthy; aim for a more concise reply while still addressing the topic.'' This flexibility can accommodate slight variations while encouraging improvements, rather than forcing computationally heavy, drastic changes. 

In contrast, for significant deviations—such as off-topic or inappropriate content—the observer uses forced feedback: ``Your response is off-topic and contains irrelevant content; provide a relevant and concise reply related to the current conversation. For example, [...].'' Here, the permissible buffer rejects the action, and the base model is required to regenerate the response until it fulfills overlay rules. This approach ensures that the model adheres strictly in critical situations, while allowing for more flexibility in less severe deviations. To facilitate timely small talk, this forced feedback is used sparingly as determined by a random factor, with a maximum limit of three regeneration attempts. 



\begin{figure}[t]
    \centering
    \includegraphics[width=\columnwidth]{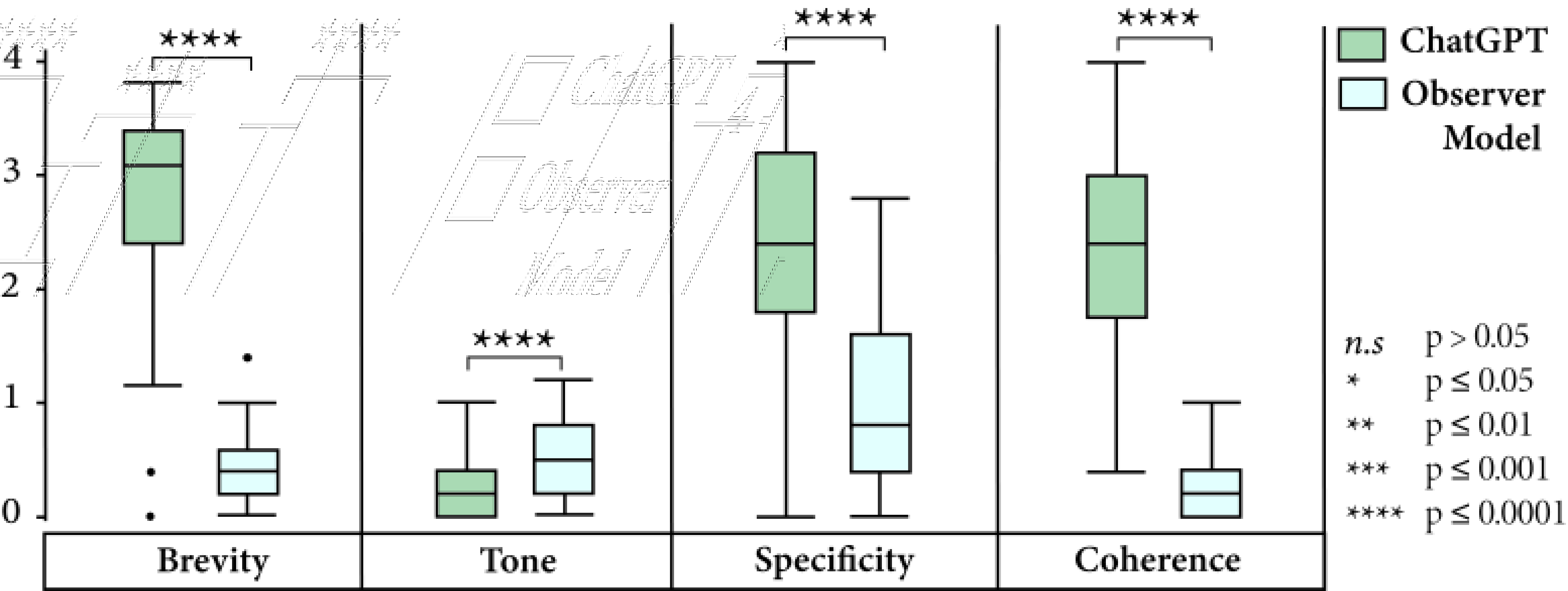}
    \caption{Evaluation of Observer v. Base Responses. The similarity of the models' small talk to that of its human users during text-based, chatbot interactions. Scores range from 0 (no difference) to 4 (highest absolute difference).}
    \label{fig:summary-observer}
\end{figure}

\subsection{Chatbot Interactions}
The participants in the baseline study engaged in 50 small talk conversations with our observer model, under the same experimental protocol. 
A total of $50$ conversations with the observer model were transcribed, yielding $499$ responses with an average of $9.98$ responses per conversation ($SD = 0.14$). Of the $250$ generated responses, $106$ ($42.4\%$) responses were flagged by the observer with implied feedback, and $14$ ($5.6\%$) responses triggered forced feedback for a total of $23$ regeneration attempts. 

We explored whether the observer's redirection was effective at improving the LLM's small talk behavior. To compare the responses of ChatGPT-3.5 (base model) in the baseline study to that with the observer-enabled system, we calculated the ``human-likeness'' of generated responses as described in the baseline along the four small talk criteria. 

The Wilcoxon method with Holm-corrected significances indicates that the observer responses were significantly more human-like in that they were more concise ($Z = -8.17$, $p \leq 0.0001$), positive ($Z = 4.53$, $p \leq 0.0001$), less specific ($Z = -6.76$, $p \leq 0.0001$), and more thematically coherent ($Z = 4.53$, $p \leq 0.0001$) than the responses of the base system. 
Furthermore, a Brown-Forsythe test on the sum of differences across small talk criteria indicates significantly less variability in human-likeness for the observer model than the base model ($F' = 15.47$, $p \leq 0.0001$). As summarized in Fig. \ref{fig:summary-observer}, the observer responses were more human-like across the criteria than the responses of the base model.

\subsection{Robot Interactions} 
Agents should have the ability to engage effectively not only in virtual, text-based interactions but also in real-world, dynamic scenarios with real users. Hence, we developed an observer-enabled robot to explore how well the system navigates the nuances of novel, face-to-face interactions. 

\begin{figure}[t]
    \centering
    \includegraphics[width=\columnwidth]{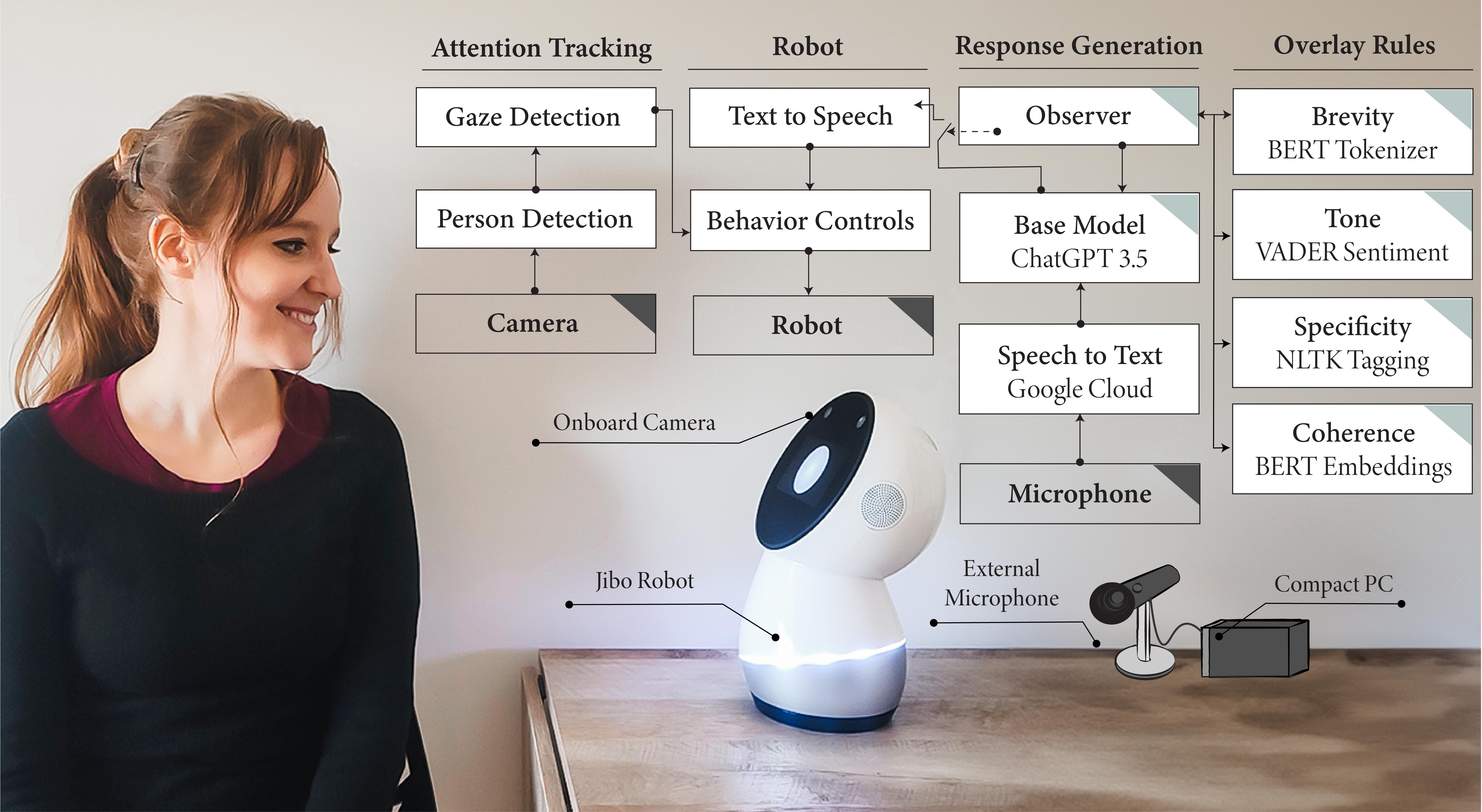}
    \caption{The observer-enabled robot engaged in naturalistic, small talk with users, fostered rapport, enhanced user comfort, and created more seamless interactions.}
    \label{fig:robot}
\end{figure}

We used the Jibo robot which stands 11 inches tall and has 3 full-revolute axes designed for 360-degree movement. We coordinated personified behaviors such as naturalistic gaze and body movement with Jibo's onboard capabilities. 
Additionally, we implement a modular software architecture within the ROS framework to allow for components of the small talk system to be fully autonomous (Fig. \ref{fig:robot}). 

A within-subjects case study was conducted where 25 volunteer participants, 15 men and 10 women, ages 19 to 45 ($M = 25.2$, $SD = 7.4$), interacted with the base-only and observer-enabled system for three conversations each. Each conversation spanned a minimum of eight turns, and the order in which participants interacted with the two models was randomized. This protocol yielded 150 conversations of 1725 responses in $\approx16.8$ hours of interaction, $40.5$ minutes ($SD = 10.2$) per participant. Following interactions with each model, participants provided open-ended feedback. We then conducted an informal thematic analysis and participant feedback was ultimately grouped into two primary themes. 

\emph{Response Content}. 21 participants expressed dissatisfaction with the base model's responses, noting its overly assistive and verbose tendencies, which led to conversations described as ``rambling'', ``dry'', and ``like speaking to a wall.'' This sentiment was echoed by $P_{25}$, who expressed frustration with the model's tendency to prioritize assistance over engaging in genuine conversation, stating, ``Even when I spoke about my own interests, it only cared about giving me help like I was a child always in need of help...'' On the other hand, in the observer condition, 23 participants remarked on how ``relevant,'' ``human-like,'' and ``natural'' were the robot's responses. For example, $P_{2}$ stated that the robot, ``engaged in small talk better than most of my friends would.'' 

\emph{Embodied Form}. 13 participants described the impact of the physical robot form on the quality of conversation. The feedback was mostly positive, highlighting that Jibo's ``animated'' and ``life-like'' movements made it ``more than a toy'' across conditions. Yet, three participants remarked on a lack of personality: ``[I]t's a bit misleading that it has a body and eyes and life-like movements but doesn't have a personality or experiences to share'' ($P_{14}$).

This exploratory study aimed to reveal users' broad perceptions of the system, demonstrating that good small talk behavior is inherently emergent and highlighting the success of the observer-enabled system. 

\section{Discussion}

Building on robotic action selection techniques, we introduced the grounded observer as a framework for aligning foundation models with desired outcomes in socially sensitive domains. Our research demonstrates this approach's usefulness by developing agents capable of seamless, contextually relevant casual conversation. In our exploratory studies, we identified gaps in existing LLMs' small talk capabilities and then enhanced a base LLM with an observer. This enhancement significantly improved the LLM's ability to follow small talk conventions, leading to more engaging and socially appropriate interactions in both virtual text-based chats and spontaneous face-to-face conversations.



While the design and internal representation of different models and robotic platforms may vary, the concept of enabling an agent to observe its own compliance goes beyond specific implementations like GPT-3.5 or Jibo. Future research should explore how the grounded observer can generalize across various platforms and behavioral contexts.

For example, the increasing use of academic tutoring systems \cite{lin2023artificial} introduces unique social risks \cite{fischer2013effects}, such as the potential for an agent to provide feedback that is overly harsh, too lenient, or even misleading, which could negatively impact students' learning and self-esteem. To mitigate these risks, overlay rules grounded in pedagogical principles can be developed \cite{price2010feedback}, ensuring feedback remains supportive, specific, and tailored to the student's progress. These rules are analogous to the small talk criteria established in our study. An observer-enabled tutoring agent could dynamically adjust its feedback to foster a positive and effective learning environment, while minimizing socially inappropriate responses.

The grounded observer framework offers significant advantages in scalability and structure, but challenges remain in the design and implementation of overlay rules. Accurately capturing nuanced behaviors is critical, as misaligned rules can lead to ineffective or inappropriate responses. Developing systematic methods for refining these rules---such as inferring rules from datasets, red-team testing \cite{hong2024curiosity}, or other methodologies \cite{bommasani2021opportunities}---is essential. Additionally, synthesizing effective overlay directives remains more art than science, underscoring the need for quantitative methods to evaluate the quality of generated prompts and to create reliable templates for observer-generated behavior. This could manifest, for example, as designing overlays for the observer's own behavior, essentially embedding quality evaluations into the agent itself. 

In all, the grounded observer framework represents a step toward establishing robust guardrails for foundation models in dynamic, unstructured, and socially sensitive contexts. 

\section*{Acknowledgments}
This work was partially funded by the National Science Foundation (NSF) under grants 1955653 and 2106690, the Office of Naval Research (ONR) grant N00014-24-1-2124, and the JST Moonshot R\&D grant JPMJMS2011, Japan. Rebecca Ramnauth is supported by the NSF GRFP and the NASEM Ford Predoctoral Fellowship.


\begin{thebibliography}{42}
\providecommand{\natexlab}[1]{#1}

\bibitem[{Alshiekh et~al.(2018)Alshiekh, Bloem, Ehlers, K{\"o}nighofer, Niekum, and Topcu}]{alshiekh2018safe}
Alshiekh, M.; Bloem, R.; Ehlers, R.; K{\"o}nighofer, B.; Niekum, S.; and Topcu, U. 2018.
\newblock Safe reinforcement learning via shielding.
\newblock In \emph{Proceedings of the AAAI conference on artificial intelligence}, volume~32.

\bibitem[{Bird, Klein, and Loper(2009)}]{bird2009natural}
Bird, S.; Klein, E.; and Loper, E. 2009.
\newblock \emph{Natural language processing with Python: analyzing text with the natural language toolkit}.
\newblock " O'Reilly Media, Inc.".

\bibitem[{Biza et~al.(2021)Biza, Wang, Platt, van~de Meent, and Wong}]{biza2021action}
Biza, O.; Wang, D.; Platt, R.; van~de Meent, J.-W.; and Wong, L.~L. 2021.
\newblock Action priors for large action spaces in robotics.
\newblock \emph{arXiv preprint arXiv:2101.04178}.

\bibitem[{Bommasani et~al.(2021)Bommasani, Hudson, Adeli, Altman, Arora, von Arx, Bernstein, Bohg, Bosselut, Brunskill et~al.}]{bommasani2021opportunities}
Bommasani, R.; Hudson, D.~A.; Adeli, E.; Altman, R.; Arora, S.; von Arx, S.; Bernstein, M.~S.; Bohg, J.; Bosselut, A.; Brunskill, E.; et~al. 2021.
\newblock On the opportunities and risks of foundation models.
\newblock \emph{arXiv preprint arXiv:2108.07258}.

\bibitem[{Brawer et~al.(2023)Brawer, Ghose, Candon, Qin, Roncone, V{\'a}zquez, and Scassellati}]{brawer2023interactive}
Brawer, J.; Ghose, D.; Candon, K.; Qin, M.; Roncone, A.; V{\'a}zquez, M.; and Scassellati, B. 2023.
\newblock Interactive policy shaping for human-robot collaboration with transparent matrix overlays.
\newblock In \emph{Proceedings of the 2023 ACM/IEEE International Conference on Human-Robot Interaction}, 525--533.

\bibitem[{Brown et~al.(2020)Brown, Mann, Ryder, Subbiah, Kaplan, Dhariwal, Neelakantan, Shyam, Sastry, Askell et~al.}]{brown2020language}
Brown, T.; Mann, B.; Ryder, N.; Subbiah, M.; Kaplan, J.~D.; Dhariwal, P.; Neelakantan, A.; Shyam, P.; Sastry, G.; Askell, A.; et~al. 2020.
\newblock Language models are few-shot learners.
\newblock \emph{Advances in neural information processing systems}, 33: 1877--1901.

\bibitem[{{Capgemini}(2023)}]{capgemini_ai}
{Capgemini}. 2023.
\newblock Why Consumers Love Generative AI: Report from the Capgemini Research Institute.
\newblock Accessed: 2024-08-08.

\bibitem[{Chen et~al.(2024)Chen, Liu, Huang, Wu, Liu, Jiang, Pu, Lei, Chen, Wang et~al.}]{chen2024large}
Chen, J.; Liu, Z.; Huang, X.; Wu, C.; Liu, Q.; Jiang, G.; Pu, Y.; Lei, Y.; Chen, X.; Wang, X.; et~al. 2024.
\newblock When large language models meet personalization: Perspectives of challenges and opportunities.
\newblock \emph{World Wide Web}, 27(4): 42.

\bibitem[{Chen and Huang(2023)}]{chen2023forgetful}
Chen, J.-T.; and Huang, C.-M. 2023.
\newblock Forgetful large language models: Lessons learned from using LLMS in robot programming.
\newblock In \emph{Proceedings of the AAAI Symposium Series}, volume~2, 508--513.

\bibitem[{Coupland(2014)}]{coupland2014small}
Coupland, J. 2014.
\newblock \emph{Small talk}.
\newblock Routledge.

\bibitem[{Denny, Kumar, and Giacaman(2023)}]{denny2023conversing}
Denny, P.; Kumar, V.; and Giacaman, N. 2023.
\newblock Conversing with copilot: Exploring prompt engineering for solving cs1 problems using natural language.
\newblock In \emph{Proceedings of the 54th ACM Technical Symposium on Computer Science Education V. 1}, 1136--1142.

\bibitem[{Devlin et~al.(2018)Devlin, Chang, Lee, and Toutanova}]{devlin2018bert}
Devlin, J.; Chang, M.-W.; Lee, K.; and Toutanova, K. 2018.
\newblock BERT: Pre-training of deep bidirectional transformers for language understanding.
\newblock \emph{arXiv preprint arXiv:1810.04805}.

\bibitem[{Dong et~al.(2024)Dong, Mu, Jin, Qi, Hu, Zhao, Meng, Ruan, and Huang}]{dong2024building}
Dong, Y.; Mu, R.; Jin, G.; Qi, Y.; Hu, J.; Zhao, X.; Meng, J.; Ruan, W.; and Huang, X. 2024.
\newblock Building guardrails for large language models.
\newblock \emph{arXiv preprint arXiv:2402.01822}.

\bibitem[{Eggins and Slade(2004)}]{eggins2004analysing}
Eggins, S.; and Slade, D. 2004.
\newblock \emph{Analysing casual conversation}.
\newblock Equinox Publishing Ltd.

\bibitem[{Fischer et~al.(2013)Fischer, Lohan, Nehaniv, and Lehmann}]{fischer2013effects}
Fischer, K.; Lohan, K.~S.; Nehaniv, C.; and Lehmann, H. 2013.
\newblock Effects of different kinds of robot feedback.
\newblock In \emph{Social Robotics: 5th International Conference, ICSR 2013, Bristol, UK, October 27-29, 2013, Proceedings 5}, 260--269. Springer.

\bibitem[{Fu et~al.(2022)Fu, Gao, Zhao, Wen, and Yan}]{fu2022learning}
Fu, T.; Gao, S.; Zhao, X.; Wen, J.-r.; and Yan, R. 2022.
\newblock Learning towards conversational AI: A survey.
\newblock \emph{AI Open}, 3: 14--28.

\bibitem[{Garcez and Lamb(2023)}]{garcez2023neurosymbolic}
Garcez, A.~d.; and Lamb, L.~C. 2023.
\newblock Neurosymbolic AI: The 3 rd wave.
\newblock \emph{Artificial Intelligence Review}, 56(11): 12387--12406.

\bibitem[{Garc{\i}a and Fern{\'a}ndez(2015)}]{garcia2015comprehensive}
Garc{\i}a, J.; and Fern{\'a}ndez, F. 2015.
\newblock A comprehensive survey on safe reinforcement learning.
\newblock \emph{Journal of Machine Learning Research}, 16(1): 1437--1480.

\bibitem[{Giray(2023)}]{giray2023prompt}
Giray, L. 2023.
\newblock Prompt engineering with ChatGPT: a guide for academic writers.
\newblock \emph{Annals of biomedical engineering}, 51(12): 2629--2633.

\bibitem[{Griffith et~al.(2013)Griffith, Subramanian, Scholz, Isbell, and Thomaz}]{griffith2013policy}
Griffith, S.; Subramanian, K.; Scholz, J.; Isbell, C.~L.; and Thomaz, A.~L. 2013.
\newblock Policy shaping: Integrating human feedback with reinforcement learning.
\newblock \emph{Advances in neural information processing systems}, 26.

\bibitem[{Gu et~al.(2022)Gu, Yang, Du, Chen, Walter, Wang, and Knoll}]{gu2022review}
Gu, S.; Yang, L.; Du, Y.; Chen, G.; Walter, F.; Wang, J.; and Knoll, A. 2022.
\newblock A review of safe reinforcement learning: Methods, theory and applications.
\newblock \emph{arXiv preprint arXiv:2205.10330}.

\bibitem[{Hong et~al.(2024)Hong, Shenfeld, Wang, Chuang, Pareja, Glass, Srivastava, and Agrawal}]{hong2024curiosity}
Hong, Z.-W.; Shenfeld, I.; Wang, T.-H.; Chuang, Y.-S.; Pareja, A.; Glass, J.; Srivastava, A.; and Agrawal, P. 2024.
\newblock Curiosity-driven red-teaming for large language models.
\newblock \emph{arXiv preprint arXiv:2402.19464}.

\bibitem[{Huang et~al.(2024)Huang, Liu, Ko, Wu, Wang, Zhang, and Tang}]{huang2024selective}
Huang, Q.; Liu, X.; Ko, T.; Wu, B.; Wang, W.; Zhang, Y.; and Tang, L. 2024.
\newblock Selective Prompting Tuning for Personalized Conversations with LLMs.
\newblock \emph{arXiv preprint arXiv:2406.18187}.

\bibitem[{Hutto and Gilbert(2014)}]{hutto2014vader}
Hutto, C.; and Gilbert, E. 2014.
\newblock Vader: A parsimonious rule-based model for sentiment analysis of social media text.
\newblock In \emph{Proceedings of the International AAAI conference on Web and Social Media}, volume~8, 216--225.

\bibitem[{Laver(1981)}]{laver1981linguistic}
Laver, J. 1981.
\newblock Linguistic routines and politeness in greeting and parting.
\newblock \emph{Conversational Routine}, 289304.

\bibitem[{Leite, Martinho, and Paiva(2013)}]{leite2013social}
Leite, I.; Martinho, C.; and Paiva, A. 2013.
\newblock Social robots for long-term interaction: a survey.
\newblock \emph{International Journal of Social Robotics}, 5: 291--308.

\bibitem[{Lin, Huang, and Lu(2023)}]{lin2023artificial}
Lin, C.-C.; Huang, A.~Y.; and Lu, O.~H. 2023.
\newblock Artificial intelligence in intelligent tutoring systems toward sustainable education: a systematic review.
\newblock \emph{Smart Learning Environments}, 10(1): 41.

\bibitem[{Lyu et~al.(2024)Lyu, Zhao, Gu, Yu, Goyal, and Arora}]{lyu2024keeping}
Lyu, K.; Zhao, H.; Gu, X.; Yu, D.; Goyal, A.; and Arora, S. 2024.
\newblock Keeping llms aligned after fine-tuning: The crucial role of prompt templates.
\newblock \emph{arXiv preprint arXiv:2402.18540}.

\bibitem[{Mesk{\'o}(2023)}]{mesko2023prompt}
Mesk{\'o}, B. 2023.
\newblock Prompt engineering as an important emerging skill for medical professionals: tutorial.
\newblock \emph{Journal of medical Internet research}, 25: e50638.

\bibitem[{Naveed et~al.(2023)Naveed, Khan, Qiu, Saqib, Anwar, Usman, Akhtar, Barnes, and Mian}]{naveed2023comprehensive}
Naveed, H.; Khan, A.~U.; Qiu, S.; Saqib, M.; Anwar, S.; Usman, M.; Akhtar, N.; Barnes, N.; and Mian, A. 2023.
\newblock A comprehensive overview of large language models.
\newblock \emph{arXiv preprint arXiv:2307.06435}.

\bibitem[{Niknazar et~al.(2024)Niknazar, Haley, Ramanan, Truong, Shrinivasan, Bhowmick, Dey, Jagmohan, Maheshwari, Ponoth et~al.}]{niknazar2024building}
Niknazar, M.; Haley, P.~V.; Ramanan, L.; Truong, S.~T.; Shrinivasan, Y.; Bhowmick, A.~K.; Dey, P.; Jagmohan, A.; Maheshwari, H.; Ponoth, S.; et~al. 2024.
\newblock Building a Domain-specific Guardrail Model in Production.
\newblock \emph{arXiv preprint arXiv:2408.01452}.

\bibitem[{OpenAI(2024)}]{openai_chatgpt}
OpenAI. 2024.
\newblock OpenAI ChatGPT API.

\bibitem[{Paul(2024)}]{paul2024continually}
Paul, S.~K. 2024.
\newblock Continually Learning Planning Agent for Large Environments guided by LLMs.
\newblock In \emph{2024 IEEE Conference on Artificial Intelligence (CAI)}, 377--382. IEEE.

\bibitem[{Price et~al.(2010)Price, Handley, Millar, and O'donovan}]{price2010feedback}
Price, M.; Handley, K.; Millar, J.; and O'donovan, B. 2010.
\newblock Feedback: all that effort, but what is the effect?
\newblock \emph{Assessment \& Evaluation in Higher Education}, 35(3): 277--289.

\bibitem[{Raman et~al.(2022)Raman, Cohen, Rosen, Idrees, Paulius, and Tellex}]{raman2022planning}
Raman, S.~S.; Cohen, V.; Rosen, E.; Idrees, I.; Paulius, D.; and Tellex, S. 2022.
\newblock Planning with large language models via corrective re-prompting.
\newblock In \emph{NeurIPS 2022 Foundation Models for Decision Making Workshop}.

\bibitem[{Ramnauth, Brscic, and Scassellati(2024)}]{ramnauth_brscic_scassellati_2024}
Ramnauth, R.; Brscic, D.; and Scassellati, B. 2024.
\newblock Small Talk for Social Robots.

\bibitem[{Shieber(2004)}]{shieber2004turing}
Shieber, S.~M. 2004.
\newblock \emph{The Turing test: Verbal behavior as the hallmark of intelligence}.
\newblock MIT Press.

\bibitem[{Team et~al.(2023)Team, Anil, Borgeaud, Wu, Alayrac, Yu, Soricut, Schalkwyk, Dai, Hauth et~al.}]{team2023gemini}
Team, G.; Anil, R.; Borgeaud, S.; Wu, Y.; Alayrac, J.-B.; Yu, J.; Soricut, R.; Schalkwyk, J.; Dai, A.~M.; Hauth, A.; et~al. 2023.
\newblock Gemini: a family of highly capable multimodal models.
\newblock \emph{arXiv preprint arXiv:2312.11805}.

\bibitem[{Touvron et~al.(2023)Touvron, Martin, Stone, Albert, Almahairi, Babaei, Bashlykov, Batra, Bhargava, Bhosale et~al.}]{touvron2023llama}
Touvron, H.; Martin, L.; Stone, K.; Albert, P.; Almahairi, A.; Babaei, Y.; Bashlykov, N.; Batra, S.; Bhargava, P.; Bhosale, S.; et~al. 2023.
\newblock Llama 2: Open foundation and fine-tuned chat models.
\newblock \emph{arXiv preprint arXiv:2307.09288}.

\bibitem[{Wang et~al.(2023)Wang, Lu, Santacroce, Gong, Zhang, and Shen}]{wang2023adapting}
Wang, K.; Lu, Y.; Santacroce, M.; Gong, Y.; Zhang, C.; and Shen, Y. 2023.
\newblock Adapting llm agents through communication.
\newblock \emph{arXiv preprint arXiv:2310.01444}.

\bibitem[{White et~al.(2023)White, Fu, Hays, Sandborn, Olea, Gilbert, Elnashar, Spencer-Smith, and Schmidt}]{white2023prompt}
White, J.; Fu, Q.; Hays, S.; Sandborn, M.; Olea, C.; Gilbert, H.; Elnashar, A.; Spencer-Smith, J.; and Schmidt, D.~C. 2023.
\newblock A prompt pattern catalog to enhance prompt engineering with chatgpt.
\newblock \emph{arXiv preprint arXiv:2302.11382}.

\bibitem[{Zhou et~al.(2022)Zhou, Muresanu, Han, Paster, Pitis, Chan, and Ba}]{zhou2022large}
Zhou, Y.; Muresanu, A.~I.; Han, Z.; Paster, K.; Pitis, S.; Chan, H.; and Ba, J. 2022.
\newblock Large language models are human-level prompt engineers.
\newblock \emph{arXiv preprint arXiv:2211.01910}.

\end{thebibliography}
\end{document}